\documentclass[10pt,twocolumn,letterpaper]{article}

\usepackage[pagenumbers]{cvpr} 

\usepackage{graphicx}
\usepackage{amsmath}
\usepackage{amssymb}
\usepackage{booktabs}
\usepackage{soul}
\usepackage{marvosym}
\usepackage{pifont}
\usepackage{overpic}
\usepackage{makecell}
\usepackage{multirow, colortbl, xcolor, booktabs}
\usepackage{xcolor}
\definecolor{airforceblue}{rgb}{0.36, 0.54, 0.66}
\definecolor{airforcered}{rgb}{0.7, 0.2, 0.2}
\definecolor{americanrose}{rgb}{1.0, 0.01, 0.24}
\definecolor{amethyst}{rgb}{0.6, 0.4, 0.8}
\definecolor{awesome}{rgb}{1.0, 0.13, 0.32}
\definecolor{voc}{rgb}{1.0, 0.13, 0.32}
\definecolor{nonvoc}{rgb}{0.0, 0.5, 0.99}

\newcommand{\minititle}[1]{\vspace{0.5em}\noindent\textbf{#1}.\hspace{0.4em}}
\newcommand{\voccolor}[1]{\textcolor{voc}{#1}}
\newcommand{\nonvoccolor}[1]{\textcolor{nonvoc}{#1}}
\newlength\savedwidth
\newcommand{\whline}[1]{\noalign{\global\savedwidth\arrayrulewidth \global\arrayrulewidth #1}%
                   \hline \noalign{\global\arrayrulewidth\savedwidth}}
\newcommand{\cmark}{\ding{51}}%
\newcommand{\xmark}{\ding{55}}%

\def\vn{{\emph{voc} $\to$} \emph{nonvoc} }
\def\nv{{\emph{nonvoc} $\to$} \emph{voc} }

\usepackage[pagebackref,breaklinks,colorlinks]{hyperref}

\usepackage[capitalize]{cleveref}
\crefname{section}{Sec.}{Secs.}
\Crefname{section}{Section}{Sections}
\Crefname{table}{Table}{Tables}
\crefname{table}{Tab.}{Tabs.}

\def\cvprPaperID{1034} 
\def\confName{CVPR}
\def\confYear{2022}

\begin{document}

\title{ContrastMask: Contrastive Learning to Segment Every Thing}

\author{Xuehui Wang$^{1\dag}$, Kai Zhao$^{2}$, Ruixin Zhang$^2$, Shouhong Ding$^2$, Yan Wang$^3$, Wei Shen$^{1{(\textrm{\Letter})}}$\\
$^1$MoE Key Lab of Artificial Intelligence, AI Institute, Shanghai Jiao Tong University\\
$^2$Youtu Lab, Tencent \ \ \ \ \ \ \ $^3$Shanghai Key Lab of Multidimensional Information Processing, ECNU \\
{\tt\small \{wangxuehui, wei.shen\}@sjtu.edu.cn; \{ruixinzhang, ericshding\}@tencent.com;} \\
{\tt\small kz@kaizhao.net; ywang@cee.ecnu.edu.cn}
}

\maketitle
\let\thefootnote\relax\footnote{$^{\textrm{\dag}}$Work done during an internship at Youtu Lab, Tencent.}
\let\thefootnote\relax\footnote{$^{\textrm{\Letter}}$Corresponding Author.}
\begin{abstract}

Partially-supervised instance segmentation is a task which requests segmenting objects from novel categories via learning on limited base categories with annotated masks thus eliminating demands of heavy annotation burden. The key to addressing this task is to build an effective class-agnostic mask segmentation model. Unlike previous methods that learn such models only on base categories, in this paper, we propose a new method, named ContrastMask, which learns a mask segmentation model on both base and novel categories under a unified pixel-level contrastive learning framework. In this framework, annotated masks of base categories and pseudo masks of novel categories serve as a prior for contrastive learning, where features from the mask regions (foreground) are pulled together, and are contrasted against those from the background, and vice versa. Through this framework, feature discrimination between foreground and background is largely improved, facilitating learning of the class-agnostic mask segmentation model. Exhaustive experiments on the COCO dataset demonstrate the superiority of our method, which outperforms previous state-of-the-arts. 
\end{abstract}
\vspace{-3mm}
\section{Introduction}
\label{sec:intro}

\begin{figure}[t]
  \centering
   \begin{overpic}[width=0.95\linewidth]{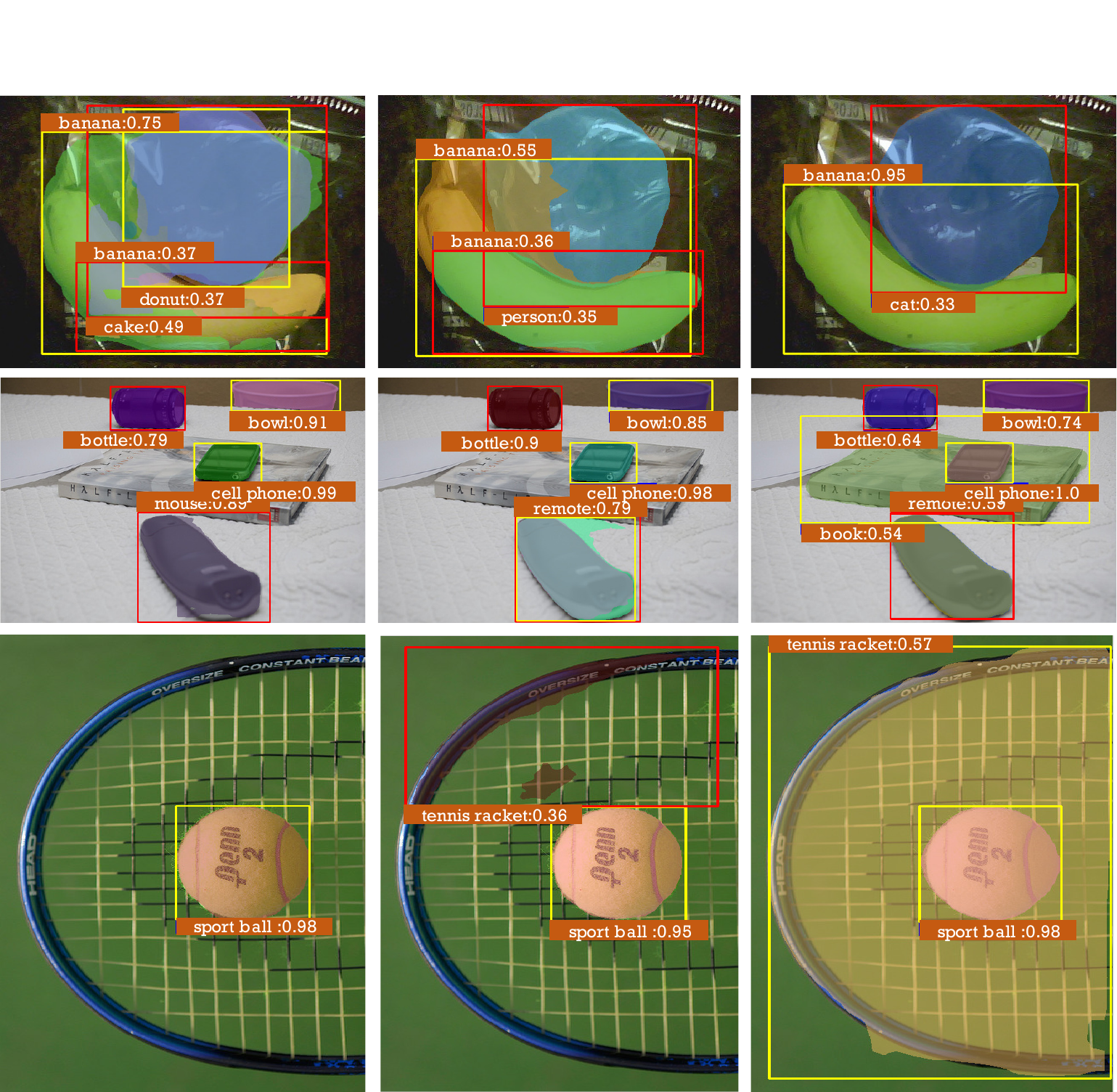}
   \put(1,91){Mask RCNN~\cite{maskrcnn}}
   \put(39,91){OPMask~\cite{opmask}}
   \put(70,91){ContrastMask}
   \end{overpic}
   \vspace{-1mm}
   \caption{Visualization results of Mask R-CNN~\cite{maskrcnn}, OPMask~\cite{opmask} and the proposed ContrastMask on \emph{novel} categories.}
   \vspace{-4mm}
   \label{fig:fig1}
\end{figure}

Instance segmentation is one of the most fundamental tasks in computer vision, which requests pixel-level prediction on holistic images and identifies each individual object. Many works~\cite{maskrcnn, condinst, htc, panet, msrcnn, polarmask, dsc, zhao2021deep} have boosted instance segmentation performance by relying on a large amount of available pixel-level annotated data. However, performing pixel-level annotation (mask annotation) is significantly burdensome, which hinders the further development of instance segmentation on massive novel categories.

Since box-level annotations are much cheaper and easier to obtain than mask annotations~\cite{boxsup}, a common way to address the aforementioned issue is to perform \emph{partially-supervised instance segmentation}~\cite{maskxrcnn, shapeprop, cpmask, shapemask}. This instance segmentation task was first proposed in the paper ``Learning to Segment Every Thing"~\cite{maskxrcnn}, where object categories are divided into two splits: \emph{base} and \emph{novel}. Both of them have box-level annotations, while only \emph{base} categories have additional mask annotations. Then the goal of this task is by taking advantage of the data of \emph{base} categories with mask annotations to generalize instance segmentation models to \emph{novel} categories. The main obstacle to achieve favorable instance segmentation performance under the partially-supervised setting is how to distinguish foreground and background within each box for an arbitrary category via learning on the data with limited annotations.

Previous methods~\cite{NIPS2015_4e4e53aa, pinheiro2016learning, maskxrcnn, shapeprop, cpmask, shapemask, opmask} addressed this task via learning a class-agnostic mask segmentation model to separate foreground and background, by capturing class-agnostic cues, such as shape bases~\cite{shapemask} and appearance commonalities~\cite{cpmask}. However, these methods learn the class-agnostic mask segmentation model only on \emph{base} categories, ignoring a large amount of training data from \emph{novel} categories, and consequently lack a bridge to transfer the segmentation capability of the mask segmentation model on \emph{base} categories to \emph{novel} categories.

To build this bridge, in this paper, we propose ContrastMask, a new partially-supervised instance segmentation method, which learns a class-agnostic mask segmentation model on both \emph{base} and \emph{novel} categories under a unified pixel-level contrastive learning framework.
In this framework, we design a new query-sharing pixel-level contrastive loss to fully exploit data from all categories. To this end, annotated masks of \emph{base} categories or pseudo masks of \emph{novel} categories computed by Class Activation Map (CAM)~\cite{cam, opmask} serve as a region prior, which indicates not only the foreground and background separation, but also shared queries, positive keys and negative keys.
Concretely, given a training image batch containing both \emph{base} categories and \emph{novel} categories, we establish two shared queries: a foreground query and a background query, which are obtained by averaging features within and outside the mask regions, including both the annotated and the pseudo masks, respectively. Then, we perform a special sampling strategy to select proper keys.
By introducing the proposed loss, we expect to pull keys within/outside the mask regions towards the foreground/background shared query and contrast it against keys outside/within the mask regions.
Finally, features learned by our pixel-level contrastive learning framework are fused into a class-agnostic mask head to perform mask segmentation.

Compared with previous methods, ContrastMask enjoys several benefits: 1) It fully exploits training data, making those from \emph{novel} categories also contribute to the optimization process of the segmentation model; 2) More importantly, it builds a bridge to transfer the segmentation capability on \emph{base} categories to \emph{novel} categories by the unified pixel-level contrastive learning framework, especially the shared queries for both \emph{base} and \emph{novel} categories, which consistently improves feature discrimination between foreground and background for both \emph{base} and \emph{novel} categories. A visualization result of comparison with other methods is shown in~\cref{fig:fig1}.

Without bells and whistles, ContrastMask surpasses all previous state-of-the-art partially-supervised instances segmentation methods on the COCO dataset~\cite{coco}, by large margins. Notably, with the ResNeXt-101-FPN~\cite{resnext, fpn} as the backbone, our method achieves 39.8 mAP for mask segmentation on \emph{novel} categories.

\section{Related Work}
\label{sec:related_work}

\textbf{Instance Segmentation.} Instance segmentation is a task that combines both object detection and semantic segmentation, \emph{i.e.}, each pixel is assigned to a specific category and an individual instance simultaneously. Mask R-CNN~\cite{maskrcnn} produced a mask for each detected bounding box by extending Faster R-CNN with a mask head. PANet~\cite{panet} improved segmentation performance by building bottom-up path augmentations and lateral connections across features of multiple levels. HTC~\cite{htc} presented interleaved execution and mask information flow and achieved considerable performance. DSC~\cite{dsc} formed a bi-directional relationship between detection and segmentation tasks, and achieved state-of-the-art performance. BMask~\cite{bmask} established a parallel head to predict the boundary of objects, which can be fused into the mask head to refine segmentation results. BCNet~\cite{bcnet} adopted bilayer GCN and self-attention to regress the object contour and instance masks, respectively. In addition to these two-stage methods, one-stage methods, such as CondInst~\cite{condinst}, BlendMask~\cite{blendmask}, SOLO\cite{solo}, SOLOv2~\cite{solo2}, YOLACT~\cite{yolact}, and FCIS~\cite{fcis}, obtained comparable performance with favorable inference speed.

\textbf{Pixel-level contrastive learning.}
Very recently, several works~\cite{densecl, xie2021propagate, Zhong_2021_ICCV, Alonso_2021_ICCV, pixel_contrast_1, pixel_contrast_2, pixel_contrast_3} have been proposed to perform pixel-level contrastive learning to remedy the misalignment between the classification task and the dense prediction task. However, these methods and ours \textbf{differ in both objective and design philosophy}: Their objective is to learn general dense representations for per-pixel multi-class categorization, so they perform pixel-level instance discrimination by sampling keys from two different augmented views; While ours is to improve the foreground and background discrimination, so we perform pixel-level instance discrimination by sampling keys from foreground and background areas within one image.

\textbf{Partially-supervised instance segmentation.} As the pioneer method of partially-supervised instance segmentation, Mask$^X$ R-CNN~\cite{maskxrcnn} designed a parameterized transformation function between the bounding box head and the mask head in Mask R-CNN~\cite{maskrcnn}, which enables segmenting \emph{novel} categories based on the assumption that the bounding box head encodes the embeddings of all categories.
ShapeMask~\cite{shapemask} learned shape priors from limited data with mask annotations, and expected these shape priors can generalize to \emph{novel} objects.
ShapeProp~\cite{shapeprop} exploited box supervision to learn salient regions and propagated these regions to the whole box proposals via an efficient message passing module which can generate a more accurate shape prior.
CPMask~\cite{cpmask} achieved promising performance by parsing shape commonality and appearance commonality among different categories. It claimed that sharing these commonalities can promote the generalization ability for mask prediction in a class-agnostic manner.
OPMask~\cite{opmask} employed object mask prior (OMP) to provide general concepts of foreground for mask head learning, and thus the network can concentrate on the primary objects in a region.
Very recently, Deep-MAC~\cite{deepMAC} explored the impact of mask head architectures to segmentation performance on \emph{novel} categories. It adopted much heavier architectures, such as Hourglass-52~\cite{hourglass}, for mask heads, and achieved outstanding performance. However, a lightweight mask head is always more popular in practice.

All these methods optimize their mask segmentation models only on \emph{base} categories, ignoring a large amount of data from \emph{novel} categories, and thus lack a bridge to transfer the segmentation capability on \emph{base} categories to \emph{novel} categories. We address this issue by introducing a unified contrastive learning framework for dense mask prediction, in which both \emph{base} and \emph{novel} categories contribute to mask segmentation model learning.

\begin{figure}[!t]
  \centering
   \begin{overpic}[width=\linewidth]{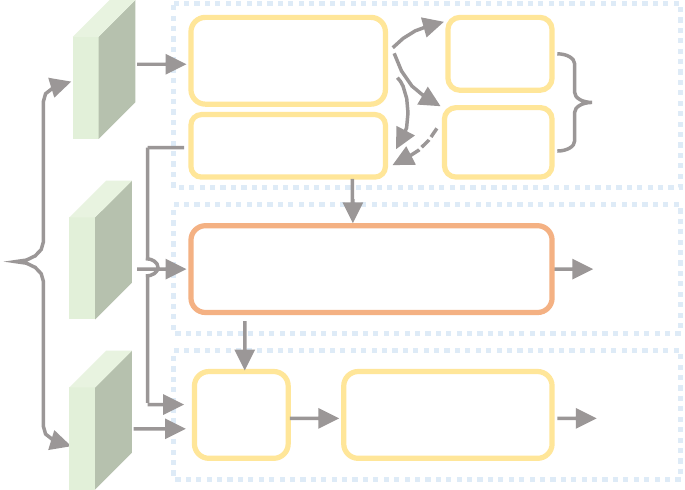}
   \put(0, 34){$\mathbf{X}$}
   \put(34,61){Box Head}
   \put(37.5,48.5){CAM}
   \put(70.5,49.5){cls}
   \put(70.5,62.5){loc}
   \put(88,55){$L_{box}$}
   \put(87,9){$L_{mask}$}
   \put(88,31){$L_{con}$}
   \put(31.5,31){Contrastive Learning Head}
   \put(38.5,20){$\mathbf{Y}$}
   \put(62,17.5){}
   \put(55.8,9.5){Mask Head}
   \put(31.5,9){Fuse}
   \put(13,60){S7}
   \put(11.5,33){S28}
   \put(11.5,9){S14}
   \end{overpic}
   \caption{The whole architecture of ContrastMask, which is built on the Mask R-CNN~\cite{maskrcnn}, with an extra contrastive learning head. ``S$n$" denotes that size of the feature map is $n\times n$. $\mathbf{X}$ and $\mathbf{Y}$ are an intput RoI feature map and its enhanced feature map, respectively.}
   \vspace{-3mm}
   \label{fig:arch}
\end{figure}
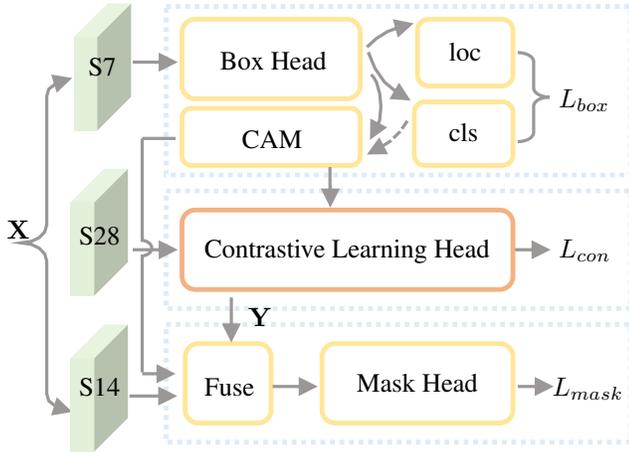

\vspace{-1mm}
\section{ContrastMask}
\label{sec:method}

We first depict the whole flowchart of the proposed ContrastMask. Then, we show how the unified pixel-level contrastive learning framework is instantiated to enhance feature discrimination between foreground and background on both \emph{base} and \emph{novel} categories. Finally, we introduce the loss functions to learn our partially-supervised instance segmentation model.

\subsection{Overview}
\vspace{-1mm}
As shown in~\cref{fig:arch}, our method, ContrastMask, is built on the classic two-stage Mask R-CNN~\cite{maskrcnn} architecture with an extra ``contrastive learning'' head, termed as CL Head, which performs unified pixel-level contrastive learning on both \emph{base} and \emph{novel} categories.
The CL Head takes an RoI feature map and a CAM generated by the box head as input.
It is supervised by our proposed pixel-level contrastive loss (~\cref{QPCL}) and outputs an enhanced feature map for the mask head.
Finally, the mask head predicts a class-agnostic segmentation map by taking a fused feature map as input.
Next, we describe the details of each component of our method.

\subsection{Contrastive Learning Head (CL Head)}

The goal of the CL Head is to increase feature discrimination between foreground and background and decrease feature dissimilarity within each region (background or foreground) for both \emph{base} and \emph{novel} categories, so that it can facilitate mask head learning. We achieve this by learning it with a new pixel-level contrastive loss.

\begin{figure*}[!t]
    \centering
     \begin{overpic}[width=\linewidth]{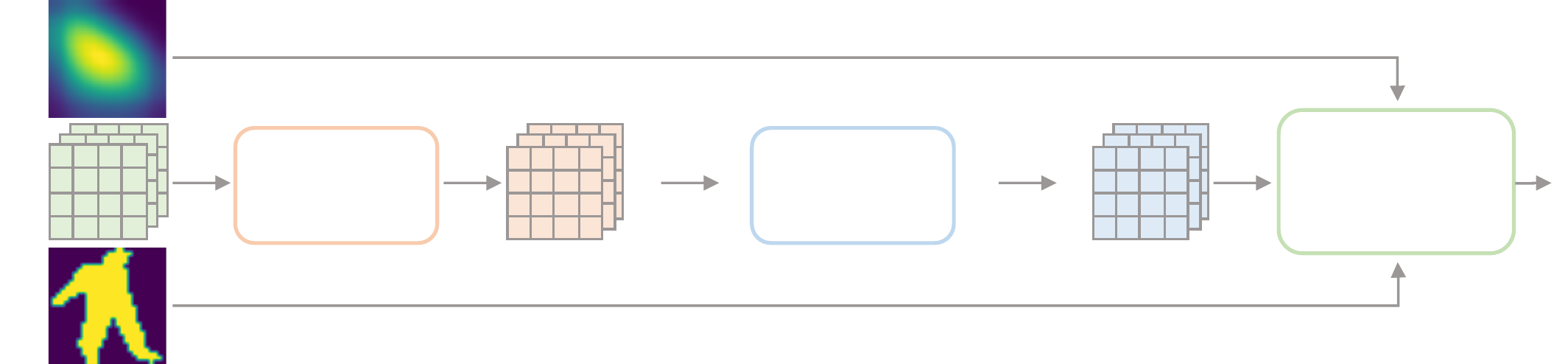}
     \put(0.8, 19){$\mathbf{A}$}
     \put(0.6, 3){$\mathbf{M}$}
     \put(0.8, 10.5){$\mathbf{X}$}
     \put(17.5, 12){\textbf{Encoder}}
     \put(19.5, 9.5){\textbf{$f(\cdot)$}}
     \put(50, 12){\textbf{Projector}}
     \put(52.5, 9.5){\textbf{$g(\cdot)$}}
     \put(35.5, 5){$\mathbf{Y}$}
     \put(41, 12.5){flatten}
     \put(62, 12.5){reshape}
     \put(73, 5){$\mathbf{Z}$}
     \put(82.5, 13.2){\textbf{Query-sharing}}
     \put(84.5, 10.9){\textbf{Pixel-level}}
     \put(87, 8.6){\textbf{$L_{con}$}}
     \put(11.5, 20.5){\textbf{Novel:} \emph{CAM (Pseudo Mask)} }
     \put(11.5, 1){\textbf{Base:} \emph{GT Mask} }
    \end{overpic}
     \caption{The flowchart of our contrastive learning head (CL Head) which consists of an encoder and a projector, supervised by a pixel-level contrastive loss. Ground-truth masks (if \emph{base}) or pseudo masks converted from CAMs (if \emph{novel}) are used to calculate the contrastive loss.}
     \label{fig:clhead}
     \vspace{-3mm}
\end{figure*}

As illustrated in ~\cref{fig:clhead}, the architecture of the contrastive learning head (CL Head) is inspired by SimCLR~\cite{simclr}, which is composed of a lightweight encoder $f(\cdot)$ and a projector $g(\cdot)$ for contrastive learning.
The encoder $f(\cdot)$ contains eight $3\times3$ Conv-ReLU layers and the projector $g(\cdot)$ is a three-layer MLP, where the last layer is not followed by a ReLU function.

Given an RoI feature map  $\mathbf{X} \in \mathbb{R}^{H\times W \times C}$ extracted by RoIAlign~\cite{maskrcnn}, where $C$, $H$ and $W$ represent channel dimension, height and width of the RoI, respectively, the CL Head feeds them into the encoder to get an enhanced feature map $\mathbf{Y}=f(\mathbf{X}) \in \mathbb{R}^{H\times W \times C}$ which will be incorporated into the mask head for mask segmentation. Next, $\mathbf{Y}$ is first flattened and then fed into the projector, which maps representations to the space where the pixel-level contrastive loss is
applied: $\mathbf{Z}=g(\mathbf{Y}) \in \mathbb{R}^{HW \times C}$. Here, after projection, the feature map $\mathbf{Z}$ is reshaped to the same size as $\mathbf{Y}$.

\begin{figure}[t]
  \centering
   \begin{overpic}[width=0.97\linewidth]{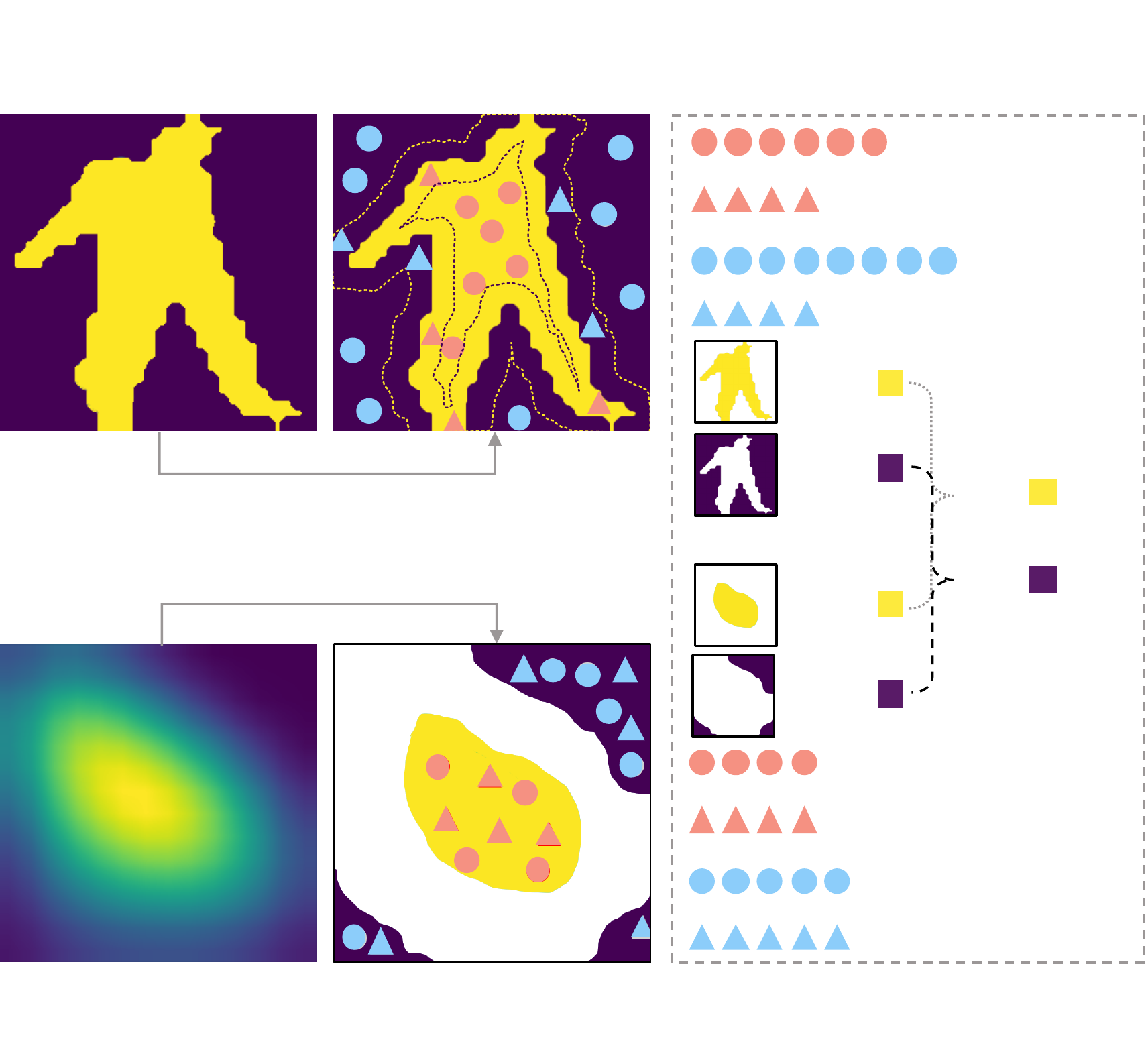}
   \put(2,88){GT Mask ($\mathbf{M}$)}
   \put(9,84){\emph{Base}}
   \put(5.5,5){CAM ($\mathbf{A}$)}
   \put(8.5,1){\emph{Novel}}
   \put(92,79){$\mathcal{K}_E^+$}
   \put(92,74){$\mathcal{K}_H^+$}
   \put(92,69){$\mathcal{K}_E^-$}
   \put(92,64){$\mathcal{K}_H^-$}
   \put(92,25.5){$\mathcal{K}_E^+$}
   \put(92,20.5){$\mathcal{K}_H^+$}
   \put(92,15.5){$\mathcal{K}_E^-$}
   \put(92,10.5){$\mathcal{K}_H^-$}
   \put(94,49){$\mathbf{q}^+$}
   \put(94,41.5){$\mathbf{q}^-$}
   \put(69.5,55){avg}
   \put(69.5,35.3){avg}
   \put(83,49){avg}
   \put(83,41.5){avg}
   \put(11,48){partition then sample}
   \put(2,42){binarize, partition then sample}
   \end{overpic}
   \caption{A schematic diagram to illustrate how to obtain queries and sample keys. For \emph{base} categories, we use ground-truth masks to do partition and extract edges to guide sampling hard keys. For \emph{novel} categories, we firstly binarize CAMs by a threshold $\delta$, then perform partition and randomly sample easy and hard keys based on partitions. The foreground query $\mathbf{q}^+$ and background query $\mathbf{q}^-$ are obtained by averaging features from corresponding partitions of a batch of object proposals.}
   \vspace{-3mm}
   \label{fig:query_key_generation}
\end{figure}
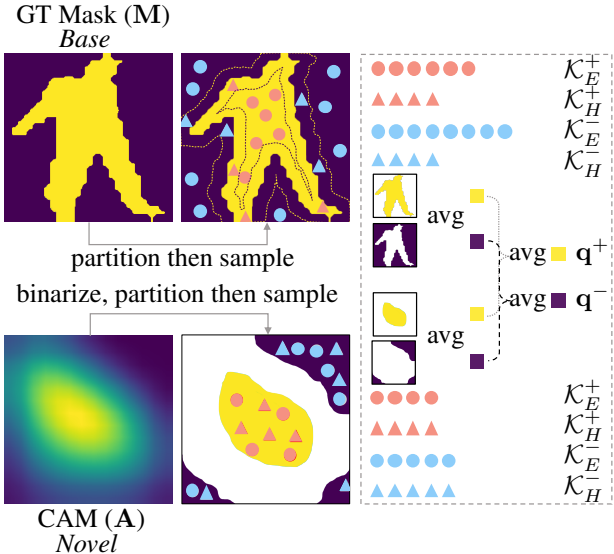

\subsection{Query-sharing Pixel-level Contrastive Loss}
\label{QPCL}

Now, we introduce our new pixel-level loss, which enables learning the mask segmentation model on both \emph{base} and \emph{novel} categories under a unified contrastive learning framework. A core design philosophy for this loss function is \emph{base} and \emph{novel} categories share two class-agnostic queries, one for foreground $\mathbf{q}^+$ and the other for background $\mathbf{q}^-$, so that a bridge is formed to transfer the segmentation capability on \emph{base} categories to \emph{novel} categories. For this reason, we name our loss function \textbf{query-sharing pixel-level contrastive loss}.

The query-sharing pixel-level contrastive loss consists of two symmetrical formulations for foreground and background, respectively. Taking foreground as an example, let $\mathcal{K}^+$ and $\mathcal{K}^-$ denote a set of foreground keys and a set of background keys, respectively. Then the query-sharing pixel-level contrastive loss for foreground is defined as:
\vspace{-3mm}

\begin{align}
\vspace{-3mm}
\label{equ:eq_0}
    &L_{\mathcal{K}^+, \mathcal{K}^-}^{\mathbf{q}^+} =
    -\frac{1}{|\mathcal{K}^+|} \sum_{\mathbf{k}^+ \in \mathcal{K}^+}   \Big[\phi(\mathbf{q}^+,\mathbf{k}^+)/\tau \\ \nonumber
    &-\log{  \big(\exp(\phi(\mathbf{q}^+,\mathbf{k}^+)/\tau) + \sum_{\mathbf{k}^- \in \mathcal{K}^-}  \exp(\phi(\mathbf{q}^+,\mathbf{k}^-)/\tau)\big)\Big]   },
\end{align}
where $\tau $ is a temperature hyper-parameter and $\phi(\cdot,\cdot)$ denotes the cosine similarity. Similarly, we can obtain the query-sharing pixel-level contrastive loss for background $L_{\mathcal{K}^-, \mathcal{K}^+}^{q^-}$.
Next, we describe the details of how to obtain the foreground and background queries $\mathbf{q}^+,\mathbf{q}^-$, as well as the foreground and background key sets $\mathcal{K}^+, \mathcal{K}^-$. We illustrate these details in~\cref{fig:query_key_generation}.

\minititle{Foreground and background partition} Given a projected feature map $\textbf{Z} \in\mathbb{R}^{H\times W\times C}$, let $\mathbf{M} \in \{0,1\}^{H\times W}$ and $\mathbf{A}\in [0,1]^{H\times W}$ be the ground-truth mask and the class-activation map (CAM) aligned with $\mathbf{Z}$, repectively. Let $\mathcal{I}$ denote the $H\times W$ spatial location lattice of feature map $\textbf{Z}$, then given a location $i\in\mathcal{I}$, we can obtain a feature vector $\mathbf{z}_i$ at location $i$ from feature map $\textbf{Z}$, and similarly the mask label $m_i$ and the CAM value $a_i$ at the location $i$ from ground-truth mask $\mathbf{M}$ and CAM $\mathbf{A}$, respectively. The whole spatial location lattice can be partitioned into two disjoint lattices: foreground location lattice $\mathcal{I}^+$ and background location lattice $\mathcal{I}^-$. For \emph{base} categories, we can achieve this partition by using the ground-truth mask: $\mathcal{I}^+=\{i\in\mathcal{I}|m_i=1\}$ and $\mathcal{I}^-=\{i\in\mathcal{I}|m_i=0\}$; While for \emph{novel} categories, as the ground-truth mask is not available, we alternatively use the CAM $\mathbf{A}$ to guide the foreground and background partition: $\mathcal{I}^+=\{i\in\mathcal{I}|a_i\ge 1-\delta\}$ and $\mathcal{I}^-=\{i\in\mathcal{I}|a_i \le \delta\}$, where $\delta=0.1$ is a small threshold and is fixed in our method.

\minititle{Query and key set generation} Let $\mathcal{I}^+_{(n)}$ and $\mathcal{I}^-_{(n)}$ be the foreground and background partitions of $n^{th}$ RoI proposal in a batch consisting of $N$ RoI proposals $\{\mathbf{Z}^{(n)}\}_{n=1}^N$, from both \emph{base} and \emph{novel} categories, respectively. The foreground and background queries $\mathbf{q}^+, \mathbf{q}^-$ are obtained by averaging features within foreground and background partitions across all proposals. Taking the foreground query $\mathbf{q}^+$ as an example, we obtain it by:
\begin{equation}
    \mathbf{q}^+ = \frac{1}{N} \sum_{n=1}^N \frac{1}{|\mathcal{I}^+_{(n)}|}\sum_{i \in \mathcal{I}^+_{(n)}} \mathbf{z}_i^{(n)}.
\end{equation}
The foreground and background key sets for a RoI proposal $\mathbf{Z}$ (here we omit the index $n$ for notation simplicity) are obtained by $\mathcal{K}^+=\{\mathbf{z_i} | i \in \S(\mathcal{I}^+,\sigma)\}$ and $\mathcal{K}^-=\{\mathbf{z_i} | i \in \S(\mathcal{I}^-,\sigma)\}$, respectively, where $\S(\cdot,\sigma)$ is a random sampling operator which samples a subset from a set randomly with a proportion ratio $\sigma$.

\minititle{Hard and easy key mining} Previous studies reveal that mining hard and easy keys is helpful to learn discriminative features by contrastive learning~\cite{understanding_contrast_loss_1,understanding_contrast_loss_2}.

For \emph{base} categories, we specify keys near an object boundary as hard keys and those far away from the boundary as easy keys. The ground-truth boundary can be obtained from the ground-truth mask easily. Let $b_i$ be the nearest boundary location to location $i$ on an RoI proposal $\mathbf{Z}$.
Then, we have the sets of hard keys and easy keys:
\begin{equation}
\begin{split}
\mathcal{K}_H &= \{ \mathbf{z}_i \ \ | \ \ i\in\S(\mathcal{I},\sigma), ||i-b_i||^2_2 \le 2 \} \\
\mathcal{K}_E &= \{ \mathbf{z}_i \ \ | \ \ i\in\S(\mathcal{I},\sigma), ||i-b_i||^2_2 > 2 \}.
\end{split}
\end{equation}

For \emph{novel} categories, since the ground-truth boundary is unavailable, we simply mine the hard and easy key sets by random sampling, \emph{i.e.}, $\mathcal{K}_H = \{ \mathbf{z}_i \ \ | \ \ i\in\S(\mathcal{I},\sigma)\} $ and $\mathcal{K}_E = \{ \mathbf{z}_i \ \ | \ \ i\in\S(\mathcal{I},\sigma)\} $.

\minititle{Instantiation of the proposed contrastive loss}Now, given an RoI proposal $\mathbf{Z}$, no matter from \emph{base} or \emph{novel} categories, we have described how to obtain foreground and background key sets from it as well as how to mine hard and easy key sets from it. Then consequently, we can obtain four types of key sets (\cref{fig:query_key_generation}) from it: 1) foreground-easy key set $\mathcal{K}^+_E$, 2) foreground-hard key set $\mathcal{K}^+_H$, 3) background-easy key set $\mathcal{K}^-_E$, and 4) background-hard key set $\mathcal{K}^-_H$. Referring to ~\cref{equ:eq_0}, our query-sharing pixel-level contrastive loss is defined as:
\vspace{-1mm}
\begin{equation}
    \begin{split}
        L_{con} &= L_{\mathcal{K}^+_E, \mathcal{K}^-_E}^{\mathbf{q}^+} +
        L_{\mathcal{K}^+_H, \mathcal{K}^-_H}^{q^+} +
                             L_{\mathcal{K}^-_E, \mathcal{K}^+_E}^{\mathbf{q}^-} +
                             L_{\mathcal{K}^-_H, \mathcal{K}^+_H}^{\mathbf{q}^-},
    \end{split}
\end{equation}
which contains four terms for the four key sets, respectively, and foreground query $\mathbf{q}^+$ and background query $\mathbf{q}^-$ are shared with keys from both \emph{base} and \emph{novel} categories.


\subsection{Class-agnostic mask head}

\begin{figure}[t]
  \centering
   \begin{overpic}[width=\linewidth]{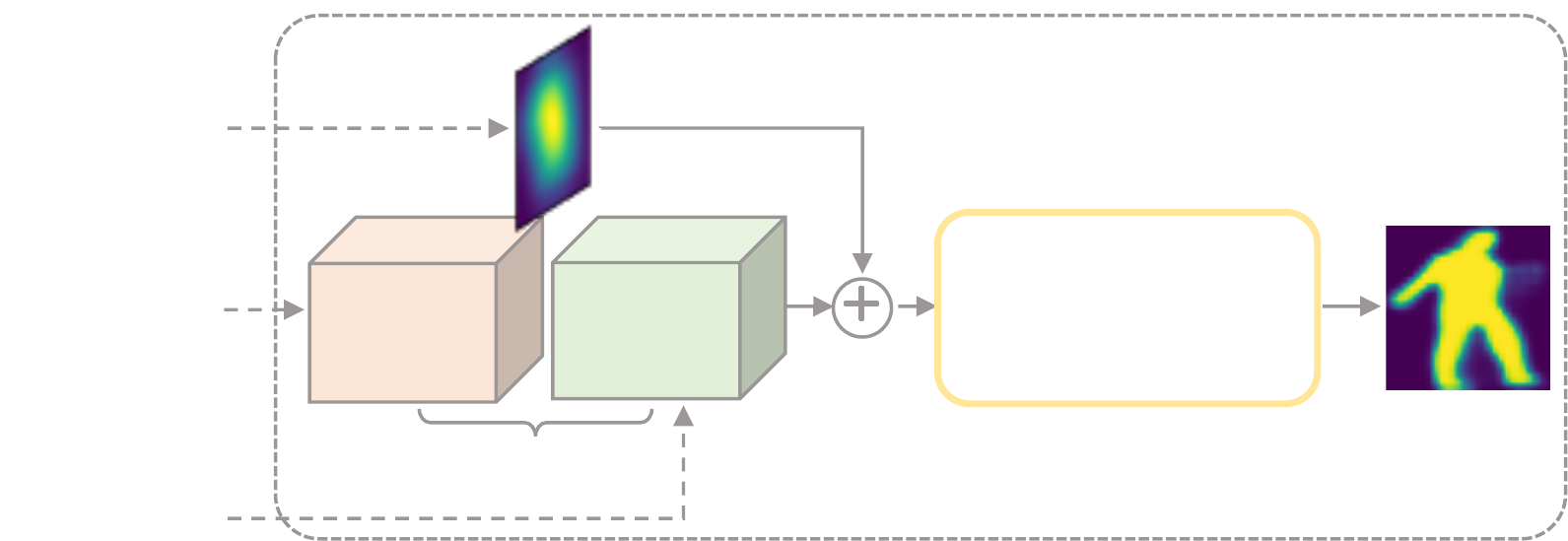}
   \put(0,4){RoIAlign}
   \put(0,17){CL Head}
   \put(0,27.5){BoxHead}
   \put(38.5,27.5){$\mathbf{A}$}
   \put(24,12.5){$\mathbf{Y}$}
   \put(40,12.5){$\mathbf{X}$}
   \put(22,4){Concatenate}
   \put(61.5,14){\textbf{Mask Head}}
   \put(86,5){$28\times28$}
   \end{overpic}
   \caption{
    The input of the class-agnostic mask head consists of enhanced featuer map $\mathbf{Y}$, RoI feature map $\mathbf{X}$ and CAM $\mathbf{A}$.
    $\oplus$ represents an element-wise addition operation.
    }
    \vspace{-3mm}
   \label{fig:class_agnostic_mask_head}
\end{figure}
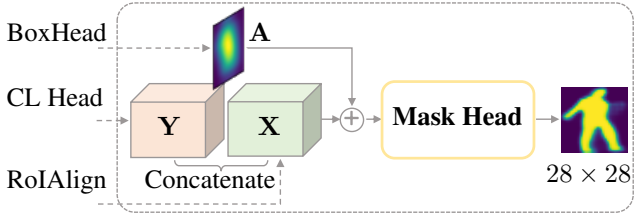

As shown in~\cref{fig:class_agnostic_mask_head}, the architecture and the corresponding loss function of the mask head is the same as those in Mask R-CNN~\cite{maskrcnn} except for three modifications: 1) Following~\cite{opmask, cpmask, shapeprop}, we change the output channels of the last convolutional layer from 80 to 1, resulting in a class-agnostic mask head. 2) We concatenate the output feature map of the CL Head with the input feature map of the mask head, which makes the input features of the mask head more distinctive and facilitates its learning. 3) We utilize the CAM~\cite{opmask} to tell the mask head which region should focus on. This can be easily implemented by adding the CAM to the input feature maps.

\subsection{Loss Function}
The overall loss function for our ContrastMask contains three terms: a box detection loss $L_{box}$, a mask segmentation loss $L_{mask}$, and a contrastive learning loss $L_{con}$. The formulations of $L_{box}$ and $L_{mask}$ are the same as those defined in Mask R-CNN~\cite{maskrcnn}:
\begin{equation}
    L = L_{box} + L_{mask} + \lambda L_{con},
    \label{eq:allloss}
\end{equation}
where $\lambda$ is a weight parameter.

\section{Experiments}
\label{sec:exps}
In this section, we first describe the experimental setup and implementation details. Then, we compare ContrastMask with state-of-the-art partially-supervised instance segmentation methods. Finally, we conduct ablation studies to show the contribution of each component in our method.

\subsection{Experimental Setup}
\label{sec:exp_setup}

Our experiments are conducted on the challenging COCO dataset\footnote{It is released under the CC-BY 4.0 license.}~\cite{coco}.
To simulate \emph{base} and \emph{novel} categories, the training set is split into two subsets.
Typically, categories presented in PASCAL VOC dataset~\cite{voc} is termed as ``\emph{voc}" and remaining categories are ``\emph{nonvoc}".
We mainly conduct experiments on these two subsets, and ``\emph{nonvoc} $\to$ \emph{voc}'' indicates that ``\emph{nonvoc}'' categories are regarded as \emph{base} and ``\emph{voc}'' as \emph{novel}, and vice versa. We use images in COCO-\emph{train2017} for training and those in COCO-\emph{val2017} for evaluation. Typical metrics for instance segmentation, \emph{i.e.}, mask AP, including mAP, AP$_{50}$, AP$_{75}$, AP$_{S}$, AP$_{M}$ and AP$_{L}$, are used for evaluation. These metrics are calculated on the \emph{novel} categories.

\minititle{Implementation details}
We implement our approach based on MMDetection\footnote{It is released under the Apache 2.0 license.}~\cite{mmdetection}. We adopt ResNet-50-FPN as the backbone for most ablation experiments and ResNet-101-FPN as the backbone for fair comparison with other methods. Typical training schedules, \emph{i.e.}, $1\times$ and $3\times$, are both employed for a fair comparison, and all ablation experiments are conducted by $1\times$ schedule for efficiency. During training, we employ SGD with momentum for optimization, and the initial learning rate is 0.02. All experiments are conducted on 8 Tesla V100 GPUs and the batch size is 16, \emph{i.e.}, 2 images per GPU. Each input image is resized to keep the rule that the long side of the image is less than 1,333 and the short side less than 800. The sampling ratio $\sigma$ is set as $\sigma=0.3$, and the temperature hyper-parameter $\tau$ (Eq.~\ref{equ:eq_0}) for easy and hard keys are set as $\tau=0.7$ and $\tau^\prime=1-\tau=0.3$, respectively. We linearly warmup the $\lambda$ of $L_{con}$ (~\cref{eq:allloss}) from 0.25 to 1.0. Besides, commonly-used augmentations such as \emph{random-flip} and multi-scale training are adopted.

\begin{table*}[!htb]
    \newcommand{\CC}[1]{\cellcolor{gray!#1}}
    \centering
    \renewcommand{\arraystretch}{1.0}
    \resizebox{\textwidth}{!}{
    \begin{tabular}{l|lcc|cccccc|cccccc}
    & & & & \multicolumn{6}{c}{\emph{nonvoc}$\rightarrow$\emph{voc}} & \multicolumn{6}{c}{\emph{voc}$\rightarrow$\emph{nonvoc}} \\
    \makecell[c]{Method} & Backbone & Sche. & Layers. & mAP & AP$_{50}$ & AP$_{75}$ & AP$_S$ & AP$_M$ & AP$_L$ & mAP & AP$_{50}$ & AP$_{75}$ & AP$_S$ & AP$_M$ & AP$_L$ \\
    \whline{1.2pt}

    Mask R-CNN (Baseline)~\cite{maskrcnn} & ResNet-50 & $1\times$ & 4  & 23.9 & 42.9 & 23.5 & 11.6 & 24.3 & 33.7 & 19.2 & 36.4 & 18.4 & 11.5 & 23.3 & 24.4  \\
    Mask$^X$ R-CNN~\cite{maskxrcnn} & ResNet-50& $1\times$ & 4           & 28.9 & 52.2 & 28.6 & 12.1 & 29.0 & 40.6 & 23.7 & 43.1 & 23.5 & 12.4 & 27.6 & 32.9  \\
    Mask GrabCut~\cite{grabcut} &ResNet-50 & $1\times$ & -               & 19.5 & 46.2 & 14.2 & 4.7  & 15.9 & 32.0 & 19.5 & 39.2 & 17.0 & 6.5  & 20.9 & 34.3  \\
    CPMask~\cite{cpmask} &  ResNet-50 &$1\times$ & 4                      & - & - & - & - & - & - & 28.8 & 46.1 & 30.6 & 12.4 & 33.1 & 43.4 \\
    ShapeProp~\cite{shapeprop} &ResNet-50 & $1\times$ & 4                 & 34.4 & 59.6 & 35.2 & 13.5 & 32.9 & 48.6 & 30.4 & 51.2 & 31.8 & 14.3 & 34.2 & 44.7  \\
    \textbf{ContrastMask (Ours)} & ResNet-50 &\textbf{$1\times$} & 4                         &\textbf{35.1} & \textbf{60.8} & \textbf{35.7} & \textbf{17.2} & \textbf{34.7} & \textbf{47.7} & \textbf{30.9} & \textbf{50.3} & \textbf{32.9} & \textbf{15.2} & \textbf{34.6} & \textbf{44.3}  \\
    \cline{1-16}
    OPMask~\cite{opmask}      &   ResNet-50    & $130k$ &  7       & 36.5 & 62.5 & 37.4 & 17.3 & 34.8 & 49.8 & 31.9 & 52.2 & 33.7 & 16.3 & 35.2 & 46.5 \\
    \textbf{ContrastMask (Ours)} & ResNet-50 &\textbf{$3\times$} & 4                         & \textbf{37.0} & \textbf{63.0} & \textbf{38.6} & \textbf{18.3} & \textbf{36.4} & \textbf{50.2} & \textbf{32.9} & \textbf{52.5} & \textbf{35.4} & \textbf{16.6} & \textbf{37.1} & \textbf{47.3} \\
    \cline{1-16}
    \textbf{ContrastMask (Ours)} & ResNeXt-50 &$3\times$ & 4                         & \textbf{37.6} & \textbf{63.8} & \textbf{38.9} & \textbf{18.1} & \textbf{36.6} & \textbf{51.3} & \textbf{33.4} & \textbf{54.2} & \textbf{35.8} & \textbf{17.7} & \textbf{37.4} & \textbf{48.5} \\

    \whline{1.2pt}
    Mask GrabCut~\cite{grabcut} &ResNet-101& $1\times$ & -               & 19.6 & 46.1 & 14.3 & 5.1  & 16.0 & 32.4 & 19.7 & 39.7 & 17.0 & 6.4  & 21.2 & 35.8  \\
    Mask$^X$ R-CNN~\cite{maskxrcnn} & ResNet-101&$1\times$ & 4           & 29.5 & 52.4 & 29.7 & 13.4 & 30.2 & 41.0 & 23.8 & 42.9 & 23.5 & 12.7 & 28.1 & 33.5  \\
    ShapeMask~\cite{shapemask}   & ResNet-101&$1\times$ & 8              & 33.3 & 56.9 & 34.3 & 17.1 & 38.1 & 45.4 & 30.2 & 49.3 & 31.5 & 16.1 & 38.2 & 28.4 \\
    ShapeProp~\cite{shapeprop} &ResNet-101 &$1\times$ & 4                 & 35.5 & 60.5 & 36.7 & 15.6 & 33.8 & 50.3 & 31.9 & 52.1 & 33.7 & 14.2 & 35.9 & 46.5  \\
    \textbf{ContrastMask (Ours)} & ResNet-101&\textbf{$1\times$} & 4                         & \textbf{36.6} & \textbf{62.2} & \textbf{37.7} & \textbf{17.5} & \textbf{36.5} & \textbf{50.1} & \textbf{32.4} & \textbf{52.1} & \textbf{34.8} & \textbf{15.2} & \textbf{36.7} & \textbf{47.3} \\
    \cline{1-16}
    ShapeMask (NAS-FPN)~\cite{shapemask}  & ResNet-101 &$3\times$ & 8     & 35.7 & 60.3 & 36.6 & 18.3 & 40.5 & 47.3 & 33.2 & 53.1 & 35.0 & 18.3 & 40.2 & 43.3 \\
    CPMask~\cite{cpmask}  & ResNet-101 &$3\times$ & 4                      & 36.8 & 60.5 & 38.6 & 17.6 & 37.1 & 51.5 & 34.0 & 53.7 & 36.5 & 18.5 & 38.9 & 47.4 \\
    OPMask~\cite{opmask}     &   ResNet-101     & $130k$    &  7       & 37.1 & 62.5 & 38.4 & 16.9 & 36.0 & 50.5 & 33.2 & 53.5 & 35.2 & 17.2 & 37.1 & 46.9 \\
    \textbf{ContrastMask (Ours)} &ResNet-101 &\textbf{$3\times$} & 4                         & \textbf{38.4} & \textbf{64.5} & \textbf{39.8} & \textbf{18.4} & \textbf{38.1} & \textbf{52.6} & \textbf{34.3} & \textbf{54.7} & \textbf{36.6} & \textbf{17.5} & \textbf{38.4} & \textbf{50.0} \\
    \cline{1-16}
    \textbf{ContrastMask (Ours)} &ResNeXt-101 &$3\times$ & 4  & \textbf{39.8} & \textbf{66.2} & \textbf{42.3} & \textbf{19.2} & \textbf{39.3} & \textbf{53.6} & \textbf{35.0} & \textbf{56.4} & \textbf{36.9} & \textbf{18.6} & \textbf{38.9} & \textbf{50.5} \\
    \end{tabular}
    }
    \caption{
        Quantitative comparisons on the challenging COCO dataset. ``\emph{nonvoc}$\to$\emph{voc}" denotes that categories in \emph{nonvoc} have the mask annotation and methods are required to be tested on \emph{voc} categories that only have box annotations, and vice versa. ``Sche." denotes the training schedule, where $1 \times$ represents for 12 epochs and \emph{130k} is a customized schedule only used in OPMask~\cite{opmask}. We use two conventional schedules, \emph{i.e.}, $1\times$ and $3\times$, for fair comparison. ``Layers." indicates the number of Conv blocks adopted in the mask head to perform mask prediction. Generally, a heavier mask head leads to better performance, which has been demonstrated in~\cite{deepMAC}. ResNeXt-50 and ResNeXt-101 indicate ``ResNeXt-50-32x4d''~\cite{resnext} and ``ResNeXt-101-64x4d''~\cite{resnext}, respectively.
    }
    \label{tab:comparisons}
\end{table*}

\subsection{Experimental Results}

We compare our method ContrastMask with recent partially-supervised instance segmentation methods, including Mask$^X$ R-CNN~\cite{maskxrcnn}, Mask GrabCut~\cite{grabcut}, ShapeMask~\cite{shapemask}, CPMask~\cite{cpmask}, ShapeProp~\cite{shapeprop} and OPMask~\cite{opmask}.

\minititle{Quantitative results}The quantitative results for \nv and \vn are shown in~\cref{tab:comparisons}. When adopting ResNet-50 as the backbone and using the $1\times$ schedule, our method surpasses the state-of-the-art method ShapeProp~\cite{shapeprop} by 0.7/0.5 mAP in \nv and \vn settings, respectively.
We also outperforms CPMask~\cite{cpmask} that uses a stronger detector, \emph{i.e.}, FCOS~\cite{fcos}, by a large margin (2.1 mAP).
In addition, we provide comparison results under the $3\times$ schedule.
Our ContrastMask (ResNet-50) achieves 37.0 mAP which even outperforms the CPMask~\cite{cpmask} (36.8 mAP) that uses ResNet-101 backbone by 0.2 mAP.
This indicates that our method fully exploits all training data and builds a bridge to transfer the segmentation capability from \emph{base} to \emph{novel}.

\begin{figure*}[t]
  \centering
   \begin{overpic}[width=0.95\linewidth]{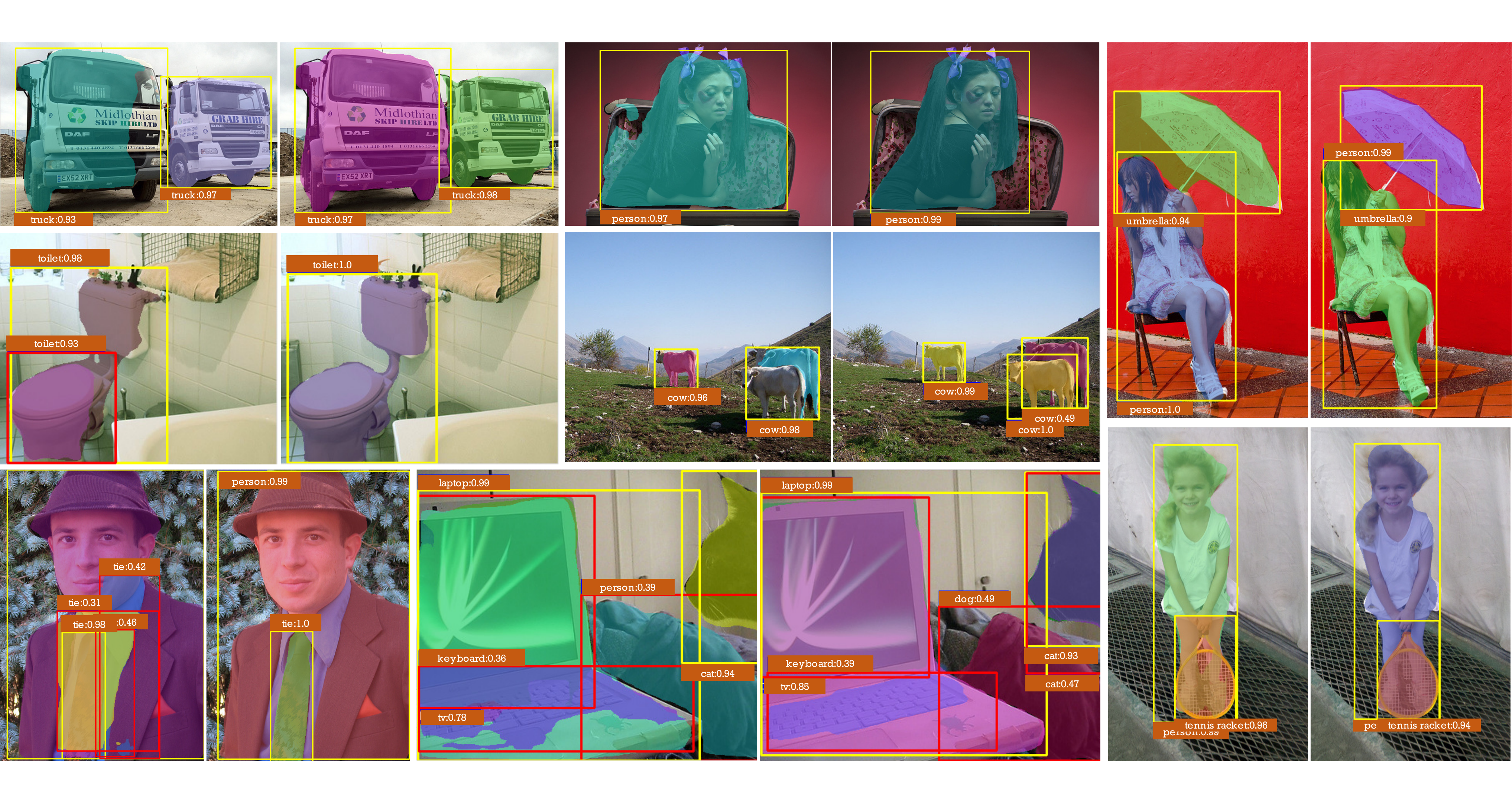}
   \put(3.5, 51){Ours $w/o$ CL}
   \put(40, 51){Ours $w/o$ CL}
   \put(74, 51){Ours $w/o$ CL}
   \put(25, 51){Ours}
   \put(62, 51){Ours}
   \put(91, 51){Ours}
   \put(7, 0.5){\emph{Base}: \voccolor{Cat, Person} \ \ \ \ \ \ \ \emph{Novel}: \nonvoccolor{Cow, Truck, Keyboard, Laptop, Tennis racket, Tie, Toilet, Umbrella}}
   \end{overpic}
   \caption{Qualitative results on COCO dataset when using \emph{voc} as training data (\emph{base}). Each group consists of two results, one is obtained by ContrastMask without CL Head (Ours $w/o$ CL) and the other is obtained by  ContrastMask (Ours). The results show that our ContrastMask performs more precisely segmentation on both \emph{base} and \emph{novel} objects benefited from the unified pixel-level contrastive learning framework conducted on all training data.}
   \label{fig:visualize_for_ps}
\end{figure*}

Our method also offers superior performance using the ResNet-101 as the backbone, \emph{e.g.}, outperforms the SOTA ShapeProp~\cite{shapeprop} by 1.1 mAP in the \nv setting.
By using the $3\times$ schedule, ContrastMask (ResNet-101) achieves new SOTA performance of 38.4/34.3 mAP in the \nv and \vn settings.
It outperforms CPMask~\cite{cpmask} and ShapeMask~\cite{shapemask} by 1.6/2.7 mAP, respectively, in the \nv setting. Note that ShapeMask~\cite{shapemask} adopts enhanced NAS-FPN~\cite{nasfpn} as the feature enhancement module to utilize multi-scale features.

%
We notice that the results of OPMask~\cite{opmask} are reported by adopting a heavier mask head, \emph{i.e.}, 7 Conv layers, and a different training schedule, \emph{i.e.}, $130k$ training iterations. We kindly refer readers to its arXiv version~\cite{opmaskarxiv} (v1) for more comparison (They reported their result under the $3\times$ schedule). Even OPMask adopts a heavier mask head, our ContrastMask still outperforms it. In addition, we also provide stronger results by using ResNeXt~\cite{resnext} backbones under the $3\times$ schedule to show the potential of our method.

\begin{figure}[!t]
  \centering
   \begin{overpic}[width=\linewidth]{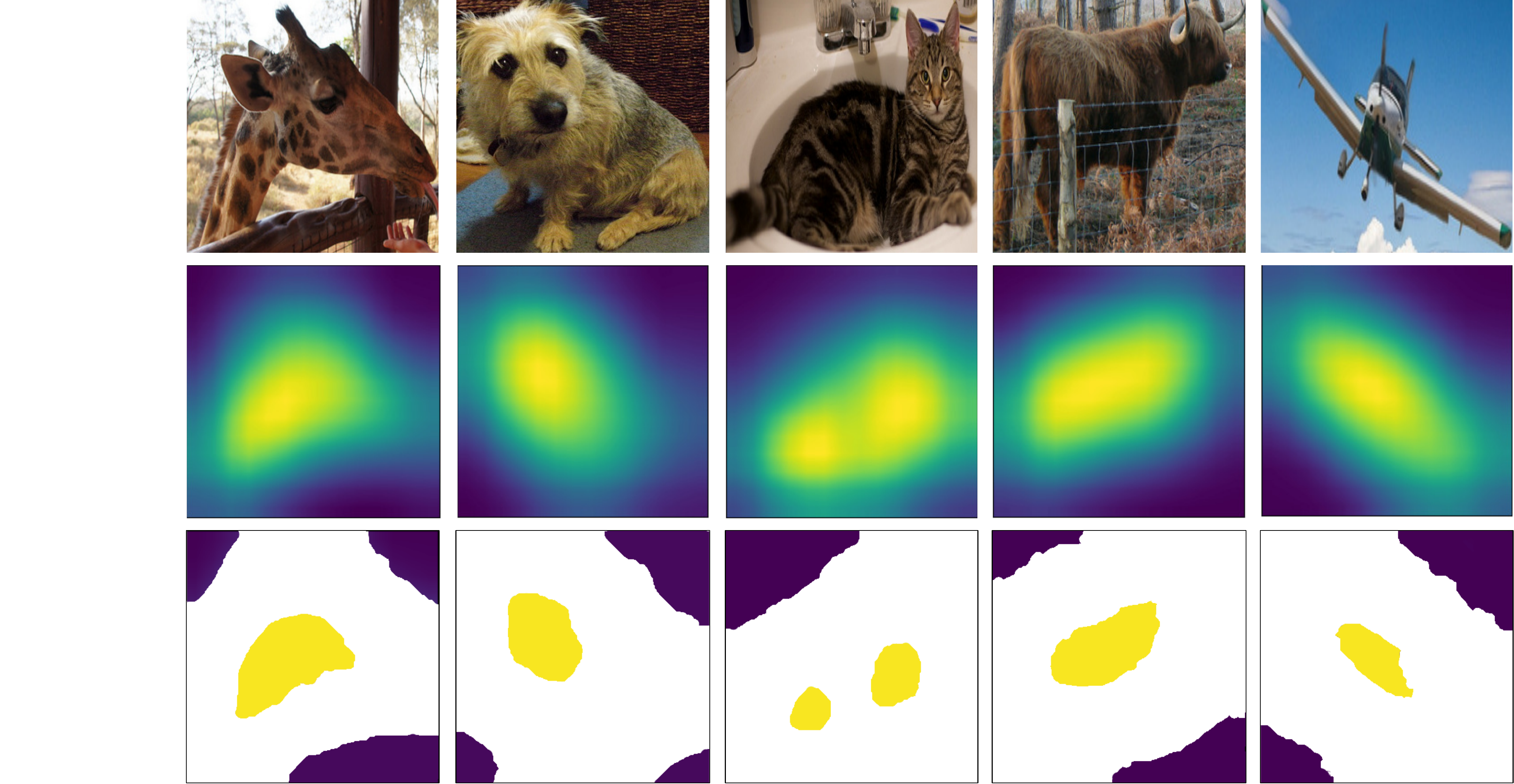}
   \put(0,42){Image}
   \put(0,25.5){CAM}
   \put(0,9){Pseudo}
   \put(0,5.5){Mask}
   \end{overpic}
   \caption{
    The visualization of CAMs and pseudo masks for some examples.
    The yellow and dark purple areas in pseudo masks denote for the foreground and background partitions, respectively.
    }
   \label{fig:pseudo_mask_supp}
\end{figure}
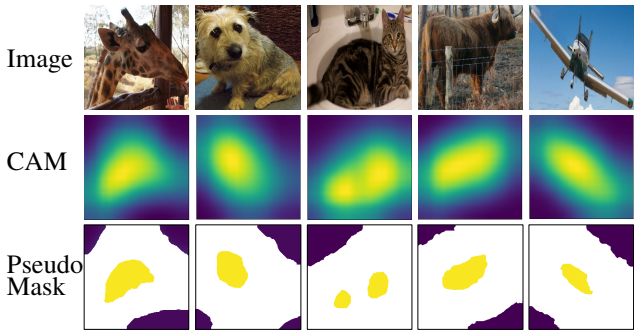

\minititle{Qualitative results}Here, we visualize some example segmentation results of our method under two situations: with and without CL Head.
We employ mask annotations from the \emph{voc} subset to train our model. In~\cref{fig:visualize_for_ps}, we show some samples from COCO-\emph{val2017} dataset, including \emph{voc} (\emph{base}) and \emph{nonvoc} (\emph{novel}) categories. Our ContrastMask represents great capability to segment both of \emph{base} and \emph{novel} objects accurately. Even if objects are small and the background is clutter, our method still performs well.
In addition, we also visualize the CAMs and corresponding pseudo masks generated by our method of some examples in~\cref{fig:pseudo_mask_supp}. These pseudo masks are used as the prior to obtain the foreground and background queries and sample keys for \emph{novel} categories.

\subsection{Ablation Study}
\label{sec: ablation_study}
We conduct ablation studies to verify different designs of the components in our ContrastMask. Unless otherwise specified, we do ablations in the \nv setting. All results are reported on \emph{novel} (\emph{voc}) categories.

\minititle{Effectiveness of CL Head}Referring to Fig.~\ref{fig:class_agnostic_mask_head}, the input of the mask head in our ContrastMask is composed of three signals: feature map $\mathbf{X}$ from the backbone, feature map $\mathbf{Y}$ from the CL head and class activation map $\mathbf{A}$ from the CAM module. Here, we do an ablation study to show the benefit brought by each of the inputs. Since Mask R-CNN~\cite{maskrcnn} is our baseline, We first train it in a partially-supervised manner. The result is shown in~\cref{tab:ablation_of_clhead}.
Then by involving the CAM module (CM) into the mask head, ``Baseline + CM'' obtains a  much better result, 32.3 mAP, since CAM brings a latent cue for class-agnostic mask head to clearly point out which region is the foreground area.
Furthermore, performance is boosted to 35.1 mAP after integrating the CL Head, termed as ``CL", with the baseline model plus the CAM module.
This evidences that the CL Head largely improves feature discrimination between foreground and background, and thus facilitates the learning of the class-agnostic mask segmentation model.

\begin{table}[!t]
    \centering
    \newcommand{\CC}[1]{\cellcolor{gray!#1}}
    \renewcommand{\arraystretch}{1.0}
    \resizebox{1\linewidth}{!}{
    \begin{tabular}{l|cccccc}
    Method & mAP & AP$_{50}$ & AP$_{75}$ & AP$_{S}$ & AP$_{M}$ & AP$_{L}$ \\
    \whline{1.2pt}
    Baseline   & 23.9 & 42.9 & 23.5 & 11.6 & 24.3 & 33.7 \\
    Baseline + CM              & 32.3 & 57.6 & 31.9 & 15.2 & 31.6 & 44.6 \\
    Baseline + CM +CL          & \CC{15}35.1 & \CC{15}60.8 & \CC{15}35.7 & \CC{15}17.2 & \CC{15}34.7 & \CC{15}47.7 \\
    \end{tabular}
    }
    \caption{
        \textbf{Ablation on the impact of each component.} The baseline is Mask R-CNN we built on. ``CM" denotes CAM and ``CL" represents for the CL head.
    }
    \vspace{-3mm}
    \label{tab:ablation_of_clhead}
\end{table}

\begin{table}[!t]
    \centering
    \newcommand{\CC}[1]{\cellcolor{gray!#1}}
    \renewcommand{\arraystretch}{1.0}
    \resizebox{0.98\linewidth}{!}{
    \begin{tabular}{c|cccccc}
    Architecture & mAP & AP$_{50}$ & AP$_{75}$ & AP$_{S}$ & AP$_{M}$ & AP$_{L}$ \\
    \whline{1.2pt}
    C4F2               & 34.2 & 59.8 & 34.6 & 16.5 & 33.7 & 46.4  \\		
    C8F3               & \CC{15}35.1 & \CC{15}60.8 & \CC{15}35.7 & \CC{15}17.2 & \CC{15}34.7 & \CC{15}47.7  \\	
    C12F4              & 35.0 & 61.1 & 35.0 & 17.3 & 34.8 & 47.5  \\
    \end{tabular}
    }
    \caption{
        \textbf{Ablation on the architecture of the CL head.} ``C$n$F$m$" indicates $n$ Conv-ReLU blocks in the encoder and $m$-layer MLP in the projector.
    }
    \vspace{-3mm}
    \label{tab:ablation_of_arch}
\end{table}

\minititle{Architecture of CL Head}Since the input to our CLHead is ROI features from the backbone,
    unlike other contrastive learning methods,
    our encoder is relatively simpler and consists of several
    convolutional and linear layers.
    Here we ablate the architecture of the encoder.
%
~\cref{tab:ablation_of_arch} illustrates different settings we explored. The base setting employs 4 Conv-ReLU blocks as the encoder and a two-layer MLP as the projector. After adding additional 4 Conv-ReLU blocks to the encoder and a one-layer MLP to the projector, an increase of 0.9 mAP (from 34.2 mAP to 35.1 mAP) is achieved, which explains that only 4 Conv-ReLU blocks are insufficient. When increasing the number of Conv-ReLU blocks to 12, the performance gain is limited. This indicates that adopting 8 Conv-ReLU blocks is adequate for an encoder, and more Conv-ReLU blocks bring limited benefits. Thus, we use ``C8F3" as the architecture of CL Head, considering the trade-off between efficiency and accuracy.

\minititle{Robustness of Sampling Ratio}A proportion ratio $\sigma$ is applied to determine the number of sampled keys for each key set. ~\cref{tab:ablation_of_sampling} shows the performance change by varying the proportion ratio. When $\sigma$ is too small or too large, \emph{i.e.}, $\sigma=0.1$ and $\sigma=0.6$, performance is degraded.
The reason is that a small $\sigma$ means only a few keys can be sampled and a small number of keys can not realize an accurate representation of foreground and background.
A large $\sigma$ encounters a dilemma that the rate of error keys will increase because the foreground and background partition for \emph{novel} categories are produced by a predicted and coarse CAM.
In general, a minor discrepancy arises among different $\sigma$, which demonstrates the robustness of our method to this hyper-parameter.
We attribute this characteristic to the fact that only two classes, \emph{i.e.}, foreground and background, are considered in our method, which requires a small number of keys to optimize the model.

\begin{table}[!t]
    \centering
    \newcommand{\CC}[1]{\cellcolor{gray!#1}}
    \renewcommand{\arraystretch}{1.0}
    \resizebox{1\linewidth}{!}{
    \begin{tabular}{c|cccccc}
    Sampling ratio $\sigma$ & mAP & AP$_{50}$ & AP$_{75}$ & AP$_{S}$ & AP$_{M}$ & AP$_{L}$ \\
    \whline{1.2pt}
    0.1                   & 34.4 & 60.2 & 34.3 & 16.8 & 34.4 & 46.9  \\
    0.2                   & 34.7 & 60.3 & 35.2 & 17.1 & 34.5 & 46.9  \\
    0.3                   & \CC{15}35.1 & \CC{15}60.8 & \CC{15}35.7 & \CC{15}17.2 & \CC{15}34.7 & \CC{15}47.7  \\
    0.6                   & 34.3 & 60.0 & 34.2 & 16.9 & 34.2 & 46.4 \\
    \end{tabular}
    }
    \caption{
        \textbf{Discussion on the sample ratio $\sigma$.}
    }\label{tab:ablation_of_sampling}
\end{table}

\begin{table}[!t]
    \centering
    \newcommand{\CC}[1]{\cellcolor{gray!#1}}
    \renewcommand{\arraystretch}{1.0}
    \resizebox{\linewidth}{!}{
    \begin{tabular}{c|ccccccc}
    Temperature $\tau$  & mAP & AP$_{50}$ & AP$_{75}$ & AP$_{S}$ & AP$_{M}$ & AP$_{L}$ \\
    \whline{1.2pt}
    0.1                & 34.4 & 60.4 & 35.0 & 16.7 & 34.1 & 46.9  \\
    0.7               & \CC{15}35.1 & \CC{15}60.8 & \CC{15}35.7 & \CC{15}17.2 & \CC{15}34.7 & \CC{15}47.7  \\
    0.9               & 34.0 & 60.2 & 33.7 & 16.8 & 33.4 & 46.4  \\
    \end{tabular}
    }
    \caption{
        \textbf{Discussion on the temperature hyper-parameter.} we apply $\tau$ and $\tau^{\prime}=1-\tau$ for easy and hard keys, respectively.
    }
    \vspace{-2mm}
    \label{tab:ablation_of_tau}
\end{table}

\minititle{Temperature hyper-parameter}
We apply $\tau$ to easy keys and $\tau^{\prime}=1-\tau$ to hard keys when computing our contrastive loss. From~\cref{tab:ablation_of_tau}, we notice that a very small $\tau$ is unsuitable for easy or hard keys, which leads to performance degradation. This can be explained from a perspective~\cite{understanding_contrast_loss_1} that only a few negative keys near the query are focused when using a small $\tau$, \emph{i.e.}, $\tau=0.1$. However, we expect more negative keys can be pushed away. Thus, we set $\tau=0.7$ for easy keys and $\tau^{\prime}=1-\tau=0.3$ for hard keys.

\minititle{Supervisions for our contrastive learning}In this study, we guide our query-sharing pixel-level contrastive learning by three different types of supervisions, \emph{i.e.}, only \emph{base}, only \emph{novel} and \emph{all}.
As shown in~\cref{tab:ablation_of_supervision}, both only using \emph{base} categories and only using \emph{novel} categories to contribute in loss calculation lead to obvious performance drops, 1.6 mAP and 1.7 mAP respectively, compared with adopting \emph{all} categories.
This demonstrates that involving training data from all categories is important to learn a segmentation model with good generalization capability between \emph{base} and \emph{novel} categories.

\minititle{Necessity of query-sharing}We ablate this experiment to validate the influence of the query-sharing strategy.
In ~\cref{tab:ablation_of_share}, ``\xmark" means that we obtain different query $\mathbf{q}$ for different proposal, and thus the pixel-level contrastive loss is calculated for each proposal individually.
It achieves worse performance compared with ``\cmark", which indicates that the query-sharing strategy is essential for the proposed unified pixel-level contrastive learning framework.

\begin{table}[!t]
    \centering
    \newcommand{\CC}[1]{\cellcolor{gray!#1}}
    \renewcommand{\arraystretch}{1.0}
    \resizebox{1\linewidth}{!}{
    \begin{tabular}{c|cccccc}
    Supervision & mAP & AP$_{50}$ & AP$_{75}$ & AP$_{S}$ & AP$_{M}$ & AP$_{L}$ \\
    \whline{1.2pt}
    \emph{base}               & 33.5 & 58.4 & 33.9 & 15.9 & 33.3 & 45.3  \\		
    \emph{novel}             & 33.4 & 58.0 & 34.2 & 15.8 & 33.1 & 45.8  \\	
    \emph{all}                & \CC{15}35.1 & \CC{15}60.8 & \CC{15}35.7 & \CC{15}17.2 & \CC{15}34.7 & \CC{15}47.7  \\
    \end{tabular}
    }
    \caption{
        \textbf{Ablation on different supervision for our contrastive learning head.} ``\emph{base}", ``\emph{novel}" and ``\emph{all}" denote that only \emph{base} categories, only ``\emph{novel}" categories and all categories are considered when calculating our contrastive loss, respectively.
    }\label{tab:ablation_of_supervision}
\end{table}

\begin{table}[!t]
    \centering
    \newcommand{\CC}[1]{\cellcolor{gray!#1}}
    \renewcommand{\arraystretch}{1.0}
    \resizebox{1\linewidth}{!}{
    \begin{tabular}{c|cccccc}
    Query-Sharing & mAP & AP$_{50}$ & AP$_{75}$ & AP$_{S}$ & AP$_{M}$ & AP$_{L}$ \\
    \whline{1.2pt}
    \xmark     & 32.7 & 56.9 & 33.1 & 15.7 & 32.0 & 44.7  \\
    \cmark     & \CC{15}35.1 & \CC{15}60.8 & \CC{15}35.7 & \CC{15}17.2 & \CC{15}34.7 & \CC{15}47.7  \\		
	
    \end{tabular}
    }
    \vspace{-1mm}
    \caption{
        \textbf{Ablation on the necessity of query-sharing}.
    }
    \vspace{-2mm}
    \label{tab:ablation_of_share}
\end{table}

\vspace{-1mm}
\section{Limitations and Discussions}
\label{sec:limitation}
\vspace{-1mm}
\minititle{Limitations}Since pseudo masks converted from CAMs are not accurate, the foreground and background partitions for \emph{novel} categories are not guaranteed to be correct, which inevitably damages segmentation performance. If ground-truth masks for \emph{novel} categories are available for sampling keys, an improvement about 1.4 mAP can be further achieved on the \vn setting. There are two ways to approach this upper bound: 1) Utilizing stronger techniques to produce more precise pseudo masks. 2) Providing scribble or point annotations for \emph{novel} categories, which are also cheaper than mask annotations.

\minititle{Possible application scenarios}Our method is a kind of label-efficient learning method, which is known as the key to extending AI algorithms to real-world applications.
%
We give two examples for reference.
1) In autonomous driving, when encountering an unknown scene where new objects are not involved in existing training set, we can annotate these objects quickly with only box annotations and then combine them with existing training set.
Then, a more accurate and robust model can be re-trained.
%
2) Suppose that a shopping mall wants to build an automatic alert system to monitor whether the fire passage is blocked by some objects. This requires training a scene parsing model.
There exist familiar objects (\emph{e.g.}, person) that have abundant annotations in public datasets or internal datasets of a company but also unusual objects (\emph{e.g.}, wheelbarrow) that do not have mask annotations.
We can apply our partially-supervised method here to reduce annotation burden.

\minititle{Relation to a teacher-student model}One may argue that our method looks like a teacher-student model, and thus the low quality of CAM may lead to negative impacts on the segmentation performance. But we think this is a kind of misunderstanding.
First, we use CAM as a prior to form query-pos/-neg pairs for CL, rather than a teacher supervision. Note that as an unsupervised/weakly-supervised learning framework, CL requires some priors, even though the priors are not always correct, to determine pos/neg keys, \emph{e.g.}, instance discrimination (MoCo~\cite{moco}, CVPR20) and color consistency (DenseCL~\cite{densecl}, CVPR21).
Second, although sometimes CAMs are incorrect, the influence is limited
(less than 1.4 mAP compared with using GTs for \emph{novel} as the prior). This is benefited from that we employ two strategies to enhance the robustness of the CL head:
1) We adopt the query-sharing strategy which forms a query based on both \emph{base} and \emph{novel} data in a batch.
2) We only consider high confidence areas of CAMs as the aforementioned prior, which diminishes the impact of errors.

\vspace{-1mm}
\section{Conclusion}
\label{sec:conclusion}
We developed an effective method for partially-supervised instance segmentation,
named as ContrastMask, which introduces a unified pixel-level contrastive learning framework to learn a segmentation model on both \emph{base} and \emph{novel} categories. 
ContrastMask utilized a query-sharing pixel-level contrastive loss to make data from \emph{novel} categories also contribute to the optimization process, and thus largely improved the feature discrimination between foreground and background areas for all categories.
These enhanced features further facilitated the learning of the class-agnostic segmentation model, resulting in a better mask segmentor.
Extensive results on the COCO dataset showed that ContrastMask consistently outperformed other methods by a large margin, achieving states-of-the-art under the partially-supervised setting.

\noindent\textbf{Acknowledgments.} This work was supported by NSFC 62176159, Natural Science Foundation of Shanghai 21ZR1432200, Shanghai Municipal Science and Technology Major Project 2021SHZDZX0102 and the Fundamental Research Funds for the Central Universities.

{\small
\bibliographystyle{ieee_fullname}
\bibliography{egbib}

\begin{thebibliography}{10}\itemsep=-1pt

\bibitem{Alonso_2021_ICCV}
I\~nigo Alonso, Alberto Sabater, David Ferstl, Luis Montesano, and Ana~C.
  Murillo.
\newblock Semi-supervised semantic segmentation with pixel-level contrastive
  learning from a class-wise memory bank.
\newblock In {\em Proceedings of the IEEE/CVF International Conference on
  Computer Vision (ICCV)}, pages 8219--8228, October 2021.

\bibitem{opmask}
David Biertimpel, Sindi Shkodrani, Anil~S. Baslamisli, and N\'ora Baka.
\newblock Prior to segment: Foreground cues for weakly annotated classes in
  partially supervised instance segmentation.
\newblock In {\em Proceedings of the IEEE/CVF International Conference on
  Computer Vision (ICCV)}, pages 2824--2833, October 2021.

\bibitem{opmaskarxiv}
David Biertimpel, Sindi Shkodrani, Anil~S. Baslamisli, and Nóra Baka.
\newblock Prior to segment: Foreground cues for weakly annotated classes in
  partially supervised instance segmentation.
\newblock {\em arXiv preprint arXiv:2011.11787v1}, 2021.

\bibitem{deepMAC}
Vighnesh Birodkar, Zhichao Lu, Siyang Li, Vivek Rathod, and Jonathan Huang.
\newblock The surprising impact of mask-head architecture on novel class
  segmentation.
\newblock In {\em Proceedings of the IEEE/CVF International Conference on
  Computer Vision (ICCV)}, pages 7015--7025, October 2021.

\bibitem{yolact}
Daniel Bolya, Chong Zhou, Fanyi Xiao, and Yong~Jae Lee.
\newblock Yolact: Real-time instance segmentation.
\newblock In {\em Proceedings of the IEEE/CVF International Conference on
  Computer Vision (ICCV)}, October 2019.

\bibitem{pixel_contrast_3}
Krishna Chaitanya, Ertunc Erdil, Neerav Karani, and Ender Konukoglu.
\newblock Contrastive learning of global and local features for medical image
  segmentation with limited annotations.
\newblock In H. Larochelle, M. Ranzato, R. Hadsell, M.~F. Balcan, and H. Lin,
  editors, {\em Advances in Neural Information Processing Systems (NeurIPS)},
  volume~33, pages 12546--12558. Curran Associates, Inc., 2020.

\bibitem{blendmask}
Hao Chen, Kunyang Sun, Zhi Tian, Chunhua Shen, Yongming Huang, and Youliang
  Yan.
\newblock Blendmask: Top-down meets bottom-up for instance segmentation.
\newblock In {\em Proceedings of the IEEE/CVF Conference on Computer Vision and
  Pattern Recognition (CVPR)}, June 2020.

\bibitem{htc}
Kai Chen, Jiangmiao Pang, Jiaqi Wang, Yu Xiong, Xiaoxiao Li, Shuyang Sun,
  Wansen Feng, Ziwei Liu, Jianping Shi, Wanli Ouyang, et~al.
\newblock Hybrid task cascade for instance segmentation.
\newblock In {\em Proceedings of the IEEE/CVF Conference on Computer Vision and
  Pattern Recognition (CVPR)}, pages 4974--4983, 2019.

\bibitem{mmdetection}
Kai Chen, Jiaqi Wang, Jiangmiao Pang, Yuhang Cao, Yu Xiong, Xiaoxiao Li,
  Shuyang Sun, Wansen Feng, Ziwei Liu, Jiarui Xu, Zheng Zhang, Dazhi Cheng,
  Chenchen Zhu, Tianheng Cheng, Qijie Zhao, Buyu Li, Xin Lu, Rui Zhu, Yue Wu,
  Jifeng Dai, Jingdong Wang, Jianping Shi, Wanli Ouyang, Chen~Change Loy, and
  Dahua Lin.
\newblock {MMDetection}: Open mmlab detection toolbox and benchmark.
\newblock {\em arXiv preprint arXiv:1906.07155}, 2019.

\bibitem{simclr}
Ting Chen, Simon Kornblith, Mohammad Norouzi, and Geoffrey Hinton.
\newblock A simple framework for contrastive learning of visual
  representations.
\newblock In {\em Internationa Conference on Machine Learning (ICML)}, pages
  1597--1607. PMLR, 2020.

\bibitem{bmask}
Tianheng Cheng, Xinggang Wang, Lichao Huang, and Wenyu Liu.
\newblock Boundary-preserving mask r-cnn.
\newblock In {\em European Conference on Computer Vision (ECCV)}, pages
  660--676. Springer, 2020.

\bibitem{boxsup}
Jifeng Dai, Kaiming He, and Jian Sun.
\newblock Boxsup: Exploiting bounding boxes to supervise convolutional networks
  for semantic segmentation.
\newblock In {\em Proceedings of the IEEE International Conference on Computer
  Vision (ICCV)}, December 2015.

\bibitem{dsc}
Hao Ding, Siyuan Qiao, Alan Yuille, and Wei Shen.
\newblock Deeply shape-guided cascade for instance segmentation.
\newblock In {\em Proceedings of the IEEE/CVF Conference on Computer Vision and
  Pattern Recognition (CVPR)}, pages 8278--8288, June 2021.

\bibitem{voc}
M. Everingham, L. Van~Gool, C.~K.~I. Williams, J. Winn, and A. Zisserman.
\newblock The pascal visual object classes (voc) challenge.
\newblock {\em International Journal of Computer Vision (IJCV)},
  88(2):303--338, June 2010.

\bibitem{cpmask}
Qi Fan, Lei Ke, Wenjie Pei, Chi-Keung Tang, and Yu-Wing Tai.
\newblock Commonality-parsing network across shape and appearance for partially
  supervised instance segmentation.
\newblock In {\em European Conference on Computer Vision (ECCV)}, pages
  379--396. Springer, 2020.

\bibitem{nasfpn}
Golnaz Ghiasi, Tsung-Yi Lin, and Quoc~V. Le.
\newblock Nas-fpn: Learning scalable feature pyramid architecture for object
  detection.
\newblock In {\em Proceedings of the IEEE/CVF Conference on Computer Vision and
  Pattern Recognition (CVPR)}, June 2019.

\bibitem{moco}
Kaiming He, Haoqi Fan, Yuxin Wu, Saining Xie, and Ross Girshick.
\newblock Momentum contrast for unsupervised visual representation learning.
\newblock In {\em Proceedings of the IEEE/CVF Conference on Computer Vision and
  Pattern Recognition}, pages 9729--9738, 2020.

\bibitem{maskrcnn}
Kaiming He, Georgia Gkioxari, Piotr Dollar, and Ross Girshick.
\newblock Mask r-cnn.
\newblock In {\em Proceedings of the IEEE International Conference on Computer
  Vision (ICCV)}, Oct 2017.

\bibitem{maskxrcnn}
Ronghang Hu, Piotr Dollár, Kaiming He, Trevor Darrell, and Ross Girshick.
\newblock Learning to segment every thing.
\newblock In {\em Proceedings of the IEEE Conference on Computer Vision and
  Pattern Recognition (CVPR)}, June 2018.

\bibitem{msrcnn}
Zhaojin Huang, Lichao Huang, Yongchao Gong, Chang Huang, and Xinggang Wang.
\newblock Mask scoring r-cnn.
\newblock In {\em Proceedings of the IEEE/CVF Conference on Computer Vision and
  Pattern Recognition (CVPR)}, pages 6409--6418, 2019.

\bibitem{bcnet}
Lei Ke, Yu-Wing Tai, and Chi-Keung Tang.
\newblock Deep occlusion-aware instance segmentation with overlapping bilayers.
\newblock In {\em Proceedings of the IEEE/CVF Conference on Computer Vision and
  Pattern Recognition (CVPR)}, pages 4019--4028, June 2021.

\bibitem{grabcut}
A. Khoreva, R. Benenson, J. Hosang, M. Hein, and B. Schiele.
\newblock Simple does it: Weakly supervised instance and semantic segmentation.
\newblock In {\em Proceedings of the IEEE Conference on Computer Vision and
  Pattern Recognition (CVPR)}, 2017.

\bibitem{shapemask}
Weicheng Kuo, Anelia Angelova, Jitendra Malik, and Tsung-Yi Lin.
\newblock Shapemask: Learning to segment novel objects by refining shape
  priors.
\newblock In {\em Proceedings of the IEEE/CVF International Conference on
  Computer Vision (ICCV)}, October 2019.

\bibitem{fcis}
Yi Li, Haozhi Qi, Jifeng Dai, Xiangyang Ji, and Yichen Wei.
\newblock Fully convolutional instance-aware semantic segmentation.
\newblock In {\em Proceedings of the IEEE Conference on Computer Vision and
  Pattern Recognition (CVPR)}, 2017.

\bibitem{fpn}
Tsung-Yi Lin, Piotr Doll{\'a}r, Ross Girshick, Kaiming He, Bharath Hariharan,
  and Serge Belongie.
\newblock Feature pyramid networks for object detection.
\newblock In {\em Proceedings of the IEEE Conference on Computer Vision and
  Pattern Recognition (CVPR)}, pages 2117--2125, 2017.

\bibitem{coco}
Tsung-Yi Lin, Michael Maire, Serge Belongie, James Hays, Pietro Perona, Deva
  Ramanan, Piotr Doll{\'a}r, and C~Lawrence Zitnick.
\newblock Microsoft coco: Common objects in context.
\newblock In {\em European Conference on Computer Vision (ECCV)}, pages
  740--755. Springer, 2014.

\bibitem{panet}
Shu Liu, Lu Qi, Haifang Qin, Jianping Shi, and Jiaya Jia.
\newblock Path aggregation network for instance segmentation.
\newblock In {\em Proceedings of the IEEE conference on computer vision and
  pattern recognition (CVPR)}, pages 8759--8768, 2018.

\bibitem{hourglass}
Alejandro Newell, Kaiyu Yang, and Jia Deng.
\newblock Stacked hourglass networks for human pose estimation.
\newblock In {\em European Conference on Computer Vision (ECCV)}, pages
  483--499. Springer, 2016.

\bibitem{pixel_contrast_1}
Pedro~O O.~Pinheiro, Amjad Almahairi, Ryan Benmalek, Florian Golemo, and
  Aaron~C Courville.
\newblock Unsupervised learning of dense visual representations.
\newblock In H. Larochelle, M. Ranzato, R. Hadsell, M.~F. Balcan, and H. Lin,
  editors, {\em Advances in Neural Information Processing Systems (NeurIPS)},
  volume~33, pages 4489--4500. Curran Associates, Inc., 2020.

\bibitem{NIPS2015_4e4e53aa}
Pedro~O O.~Pinheiro, Ronan Collobert, and Piotr Dollar.
\newblock Learning to segment object candidates.
\newblock In C. Cortes, N. Lawrence, D. Lee, M. Sugiyama, and R. Garnett,
  editors, {\em Advances in Neural Information Processing Systems (NeurIPS)},
  volume~28. Curran Associates, Inc., 2015.

\bibitem{pinheiro2016learning}
Pedro~O Pinheiro, Tsung-Yi Lin, Ronan Collobert, and Piotr Doll{\'a}r.
\newblock Learning to refine object segments.
\newblock In {\em European Conference on Computer Vision (ECCV)}, pages 75--91.
  Springer, 2016.

\bibitem{condinst}
Zhi Tian, Chunhua Shen, and Hao Chen.
\newblock Conditional convolutions for instance segmentation.
\newblock In {\em Computer Vision--ECCV 2020: 16th European Conference,
  Glasgow, UK, August 23--28, 2020, Proceedings, Part I 16}, pages 282--298.
  Springer, 2020.

\bibitem{fcos}
Zhi Tian, Chunhua Shen, Hao Chen, and Tong He.
\newblock Fcos: Fully convolutional one-stage object detection.
\newblock In {\em Proceedings of the IEEE/CVF international conference on
  computer vision (ICCV)}, pages 9627--9636, 2019.

\bibitem{understanding_contrast_loss_1}
Feng Wang and Huaping Liu.
\newblock Understanding the behaviour of contrastive loss.
\newblock In {\em Proceedings of the IEEE/CVF Conference on Computer Vision and
  Pattern Recognition (CVPR)}, pages 2495--2504, June 2021.

\bibitem{understanding_contrast_loss_2}
Tongzhou Wang and Phillip Isola.
\newblock Understanding contrastive representation learning through alignment
  and uniformity on the hypersphere.
\newblock In {\em International Conference on Machine Learning (ICML)}, pages
  9929--9939. PMLR, 2020.

\bibitem{solo}
Xinlong Wang, Tao Kong, Chunhua Shen, Yuning Jiang, and Lei Li.
\newblock Solo: Segmenting objects by locations.
\newblock In {\em European Conference on Computer Vision (ECCV)}, pages
  649--665. Springer, 2020.

\bibitem{solo2}
Xinlong Wang, Rufeng Zhang, Tao Kong, Lei Li, and Chunhua Shen.
\newblock Solov2: Dynamic and fast instance segmentation.
\newblock In H. Larochelle, M. Ranzato, R. Hadsell, M.~F. Balcan, and H. Lin,
  editors, {\em Advances in Neural Information Processing Systems (NeurIPS)},
  volume~33, pages 17721--17732. Curran Associates, Inc., 2020.

\bibitem{densecl}
Xinlong Wang, Rufeng Zhang, Chunhua Shen, Tao Kong, and Lei Li.
\newblock Dense contrastive learning for self-supervised visual pre-training.
\newblock In {\em Proceedings of the IEEE/CVF Conference on Computer Vision and
  Pattern Recognition (CVPR)}, 2021.

\bibitem{pixel_contrast_2}
Enze Xie, Jian Ding, Wenhai Wang, Xiaohang Zhan, Hang Xu, Peize Sun, Zhenguo
  Li, and Ping Luo.
\newblock Detco: Unsupervised contrastive learning for object detection.
\newblock In {\em Proceedings of the IEEE/CVF International Conference on
  Computer Vision (ICCV)}, pages 8392--8401, 2021.

\bibitem{polarmask}
Enze Xie, Peize Sun, Xiaoge Song, Wenhai Wang, Xuebo Liu, Ding Liang, Chunhua
  Shen, and Ping Luo.
\newblock Polarmask: Single shot instance segmentation with polar
  representation.
\newblock In {\em Proceedings of the IEEE/CVF conference on computer vision and
  pattern recognition (CVPR)}, pages 12193--12202, 2020.

\bibitem{resnext}
Saining Xie, Ross Girshick, Piotr Dollar, Zhuowen Tu, and Kaiming He.
\newblock Aggregated residual transformations for deep neural networks.
\newblock In {\em Proceedings of the IEEE Conference on Computer Vision and
  Pattern Recognition (CVPR)}, July 2017.

\bibitem{xie2021propagate}
Zhenda Xie, Yutong Lin, Zheng Zhang, Yue Cao, Stephen Lin, and Han Hu.
\newblock Propagate yourself: Exploring pixel-level consistency for
  unsupervised visual representation learning.
\newblock In {\em Proceedings of the IEEE/CVF Conference on Computer Vision and
  Pattern Recognition (CVPR)}, pages 16684--16693, 2021.

\bibitem{zhao2021deep}
Kai Zhao, Qi Han, Chang bin Zhang, Jun Xu, and Ming ming Cheng.
\newblock Deep hough transform for semantic line detection.
\newblock {\em IEEE Transactions on Pattern Analysis and Machine Intelligence
  (TPAMI)}, 2021.

\bibitem{Zhong_2021_ICCV}
Yuanyi Zhong, Bodi Yuan, Hong Wu, Zhiqiang Yuan, Jian Peng, and Yu-Xiong Wang.
\newblock Pixel contrastive-consistent semi-supervised semantic segmentation.
\newblock In {\em Proceedings of the IEEE/CVF International Conference on
  Computer Vision (ICCV)}, pages 7273--7282, October 2021.

\bibitem{cam}
Bolei Zhou, Aditya Khosla, Agata Lapedriza, Aude Oliva, and Antonio Torralba.
\newblock Learning deep features for discriminative localization.
\newblock In {\em Proceedings of the IEEE Conference on Computer Vision and
  Pattern Recognition (CVPR)}, pages 2921--2929, 2016.

\bibitem{shapeprop}
Yanzhao Zhou, Xin Wang, Jianbin Jiao, Trevor Darrell, and Fisher Yu.
\newblock Learning saliency propagation for semi-supervised instance
  segmentation.
\newblock In {\em IEEE/CVF Conference on Computer Vision and Pattern
  Recognition (CVPR)}, June 2020.

\end{thebibliography}
}

\end{document}